\documentclass[double]{article}
\usepackage{times}
\usepackage{algorithm}
\usepackage{algorithmic}
\usepackage{wrapfig}
\usepackage{verbatim}
\usepackage{amsmath}
\usepackage{graphicx}
\usepackage{subcaption}
\usepackage[active]{srcltx}
\usepackage{color}
\usepackage{lineno}
\usepackage{dirtytalk}
\usepackage{amsthm}
\usepackage{authblk}
\usepackage{listings}

\usepackage{epsfig,graphicx,amsfonts}
\usepackage{amsmath,amsthm,amssymb}
\usepackage{xcolor}

\usepackage{adjustbox}
\usepackage{multirow}
\usepackage{setspace}
\usepackage{hyperref}
\usepackage{comment}

\long\def\comment #1\commentend{}

\begin{document}

\title{\Large Exploration-Exploitation Model of Moth-Inspired Olfactory Navigation}
\author{Teddy Lazebnik$^{1,2,*}$, Yiftach Golov$^{3}$, Roi Gurka$^{4}$, Ally Harari$^{3}$, Alex Liberzon$^{5}$\\ \(^1\) Department of Mathematics, Ariel University, Ariel, Israel \\ \(^2\) Department of Cancer Biology, Cancer Institute, University College London, London, UK \\  \(^3\) Department of Entomology, The Volcani Center, Israel \\ \(^4\) Department of Physics and Engineering Science, Coastal Carolina University, SC, USA \\ \(^5\) Turbulence Structure Laboratory, School of Mechanical Engineering, Tel Aviv University, Israel \\ \(*\) Corresponding author: lazebnik.teddy@gmail.com }

\date{}

\maketitle 

\begin{abstract} 
Navigation of male moths toward females during the mating search offers a unique perspective on the exploration-exploitation (EE) model in decision-making. This study uses the EE model to explain male moth pheromone-driven flight paths. We leverage wind tunnel measurements and 3D tracking using infrared cameras to gain insights into male moth behavior. During the experiments in the wind tunnel, we add disturbance to the airflow and analyze the effect of increased fluctuations on moth flights in the context of the proposed EE model. We separate the exploration and exploitation phases by applying a genetic algorithm to the dataset of moth 3D trajectories. First, we demonstrate that the exploration-to-exploitation rate (EER) increases with distance from the source of the female pheromone, which can be explained in the context of the EE model. Furthermore, our findings reveal a compelling relationship between EER and increased flow fluctuations near the pheromone source. Using the open-source pheromone plume simulation and our moth-inspired navigation model, we explain why male moths exhibit an enhanced EER as turbulence levels increase, emphasizing the agent's adaptation to dynamically changing environments. This research extends our understanding of optimal navigation strategies based on general biological EE models and supports the development of advanced, theoretically supported bio-inspired navigation algorithms. We provide important insights into the potential of bio-inspired navigation models for addressing complex decision-making challenges.  \\ 


\end{abstract}

\thispagestyle{empty}

\pagestyle{myheadings} \markboth{Draft:  \today}{Draft:  \today}
\setcounter{page}{1}

\section{Introduction}
\label{sec:introduction}
Male moths can navigate challenging and dynamic environments with complex turbulent flows, such as forests and canopies, to locate females of the same species over remarkably vast distances \cite{Elkinton1987,Brady:1989}. This phenomenon holds significance for biologists and engineers, as moths exhibit remarkable abilities to perform complex tasks despite their limited cognitive and sensory capacities. Recent evidence underscores the importance of olfactory navigation, with a growing body of research focused on this aspect, as highlighted by studies such as \cite{Bau:2015, Grunbaum2015, Baker2018}. Moths adeptly navigate using spatially and temporally local cues carried by turbulent air, utilizing chemo-receptors on their antennae for chemical sensing. Additionally, they employ visual optometry for spatial orientation, as indicated by studies like \cite{Schneider1964, Vickers2000, Slifer:1970uo}. Despite these insights, our current understanding of how the moth's cognitive or sensory mechanisms operate remains incomplete, underscoring the need for a new theoretical framework to unravel these intricate processes comprehensively. Ongoing research, exemplified by studies such as \cite{Vickers2006, Harari2011}, emphasizes the pressing requirement for advancing our theoretical understanding in this fascinating field.

While observing the navigational paths of moths, researchers noted a complex behavior characterized by the interchange of two similar yet distinct patterns: a relatively narrow "zigzagging" and a broader "side-slips" motion \cite{Kennedy1983, David1983}. This observation prompted a growing body of research aimed at capturing and explaining this behavior, employing a variety of models \cite{Tobin:1981a, Baker:1997a, Carde:1997a, Vickers:1999a, Kennedy1974, Kennedy1983, Carde1984, Baker1984, Willis1991, Mafra-Neto1996a}.

Various approaches, some of which rely on specific assumptions of probabilistic behavior, memory, olfactory signal, or previous knowledge of moths, were shown to capture the observed data and reproduce outcomes similar to those biologically observed \cite{Willis1991, Belanger1998, Carde2008, Balkovsky2002, Vergassola2007, Li2001, Li2010}. Recently, Macedo et al.~\cite{Macedo2019} developed a simulator to compare various bio-inspired and engineered strategies for chemical plume tracking. Although this simulator reasonably replicates known biological processes, it is based on the diffusion process without accounting for wind or turbulence. Consequently, it is unsuitable for exploring moth navigation with the effect of turbulence as one of the core elements of moth-inspired navigation strategy~\cite{Li2001}.

In response to this limitation, Golov et al.~\cite{Golov:2021}, based on the previous simulator of pheromone plumes, introduced a computational platform, abbreviated MothPy, that simulates the behavior of moth-like navigators, considering turbulence-driven convection and diffusion of pheromone clouds from a pulsating source. Although it provided the framework to study the problem in two-dimensional horizontal plumes, it must be extended to mimic moth navigation in 3D. Furthermore, a mathematical formalization is required to generalize these outcomes for generic olfactory navigation. 

Previous biological studies focusing on the moth navigation falsely assume that the wind direction is carrying pheromones generated by the female moth in a laminar, diffusive manner, keeping the integrity of the pheromone clouds as well as predictability over time and space \cite{Baker1985, alex_intro_3}. These studies ignored the presence of turbulence and its effect on the pheromone pockets spreading \cite{MafraNeto1994,alex_intro_1, alex_intro_2}, despite the observed influence of turbulence and wind direction fluctuations (meandering) \cite{alex_intro_4, Liberzon2017}. However, the models, including the effect of turbulent velocity fluctuations, were not validated using real-world biological data, which leaves the question of their corroboration.  

In this study, we propose an Exploration-Exploitation (EE) model for moth-inspired olfactory navigation where moths balance between \textbf{exploring} options as they search roughly speaking ``cross-wind'' for olfactory signal (physically present within clouds of pheromone) and \textbf{exploiting} after the pheromone signal was obtained and the wind direction is locally evaluated. The EE trade-off was observed in other problems in nature, and it plays a pivotal role in many decision-making processes \cite{eve_nature_1,eve_nature_2,eve_nature_3,eve_nature_5}. 
Various animal species face an EE trade-off when optimizing their chances of finding food, mates, and suitable habitats \cite{eve_nature_4}. For instance, foraging birds exhibit exploration by scouting new locations to find food while exploiting reliable food sources they have come across in the past \cite{birds_ee_1,birds_ee_2}. Similarly, specific fish species migrate to explore different environments during breeding seasons, while others stay in familiar habitats to exploit available resources consistently \cite{fish_ee_1,fish_ee_2}. Despite its general usage and ability to explain various biological phenomena, the EE model has yet to be adopted for moth olfactory navigation as far as we know. 

The contribution of this work lies in three main accomplishments. First, show that an EE-based model without memory and limited sensory data can reasonably explain moth olfactory navigation in terms of the standard deviation in the data captured by the coefficient of determination metric. Second, we show that the moth's navigation strategy and behavior are greatly affected by turbulent-like air flow fluctuations during the flight. Finally, we enhance the simulator proposed by Golov et al.~\cite{Golov:2021}, improving its ability to mimic the moth's olfactory navigation dynamics.

The rest of the paper is organized as follows. Section \ref{sec:methods} formally outlines the biological dataset, moth navigation simulator, EE model definition and evaluation, and turbulence effect on the moth navigation influence analysis. Next, Section \ref{sec:results} presents the results obtained from the analysis. Finally, Section \ref{sec:discussion} discusses the biological and engineering outcomes and proposes possible future work.  

\section{Methods}
\label{sec:methods}

\begin{figure}
    \centering
    \includegraphics[width=0.99\textwidth]{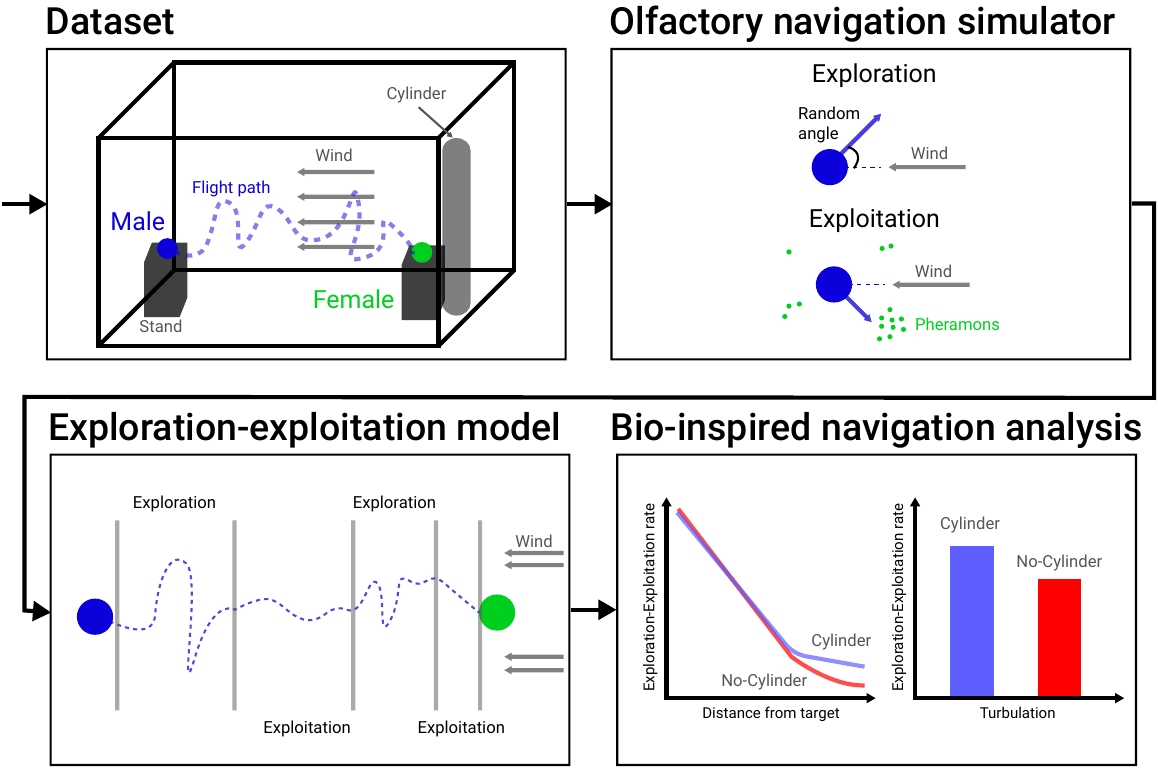}
    \caption{A schematic view of the workflow, starting with the experimental dataset, followed by the olfactory navigation simulator, exploration-exploitation-based model, and a bio-inspired analysis for flight navigation strategies. }
    \label{fig:scheme}
\end{figure}

This study is divided into four main components: I) We use moth flights in a controlled wind tunnel setting with a known wind speed and two flow conditions: undisturbed and disturbed (explained in detail herein). II) In parallel, we simulate the moth’s olfactory navigation using the extended MothPy simulator~\cite{Golov:2021}. We add several modifications to make it biologically relevant by showing the model can reasonably explain the experimental dataset. III) Using these two components, we defined and evaluated an EE-based model on a moth’s olfactory navigation. IV) Finally, we explore several bio-inspired navigation strategies and their properties using the EE-based model. Fig.~\ref{fig:scheme} provides a schematic view of the proposed study. 

\subsection{Experimental dataset}

\subsubsection{Apparatus and materials}

The pink bollworm moths were reared under laboratory control conditions at $25^\circ$C, 60\%, and 14:10 L:D at the Department of Entomology, The Volcani Center, Israel. Virgin adults were obtained by separating males and females following~\cite{dou2019pheromone}. Newly emerged adult males or females were placed in screen cages and fed 10\% sugar solution ad libitum. Three-day-old virgin males and females were used in all experiments. The experiments occurred in an open-circuit wind tunnel $(1 \times 1 \times 3$ m$^3)$ in a dark room where temperature and relative humidity were controlled at $25^\circ$ C and 60\%, respectively. The mean wind speed was set to 0.25 m/sec.

The tunnel has a built-in tracking system (Scope visual platform model Luteus-1300M-L) and an array of infrared (IR) lights outside the moths’ visible spectrum. Moth tracking CMOS cameras operating at 75 Hz were in stereoscopic viewing mode, generating a three-dimensional image field of ($ 1 \times 1 \times 2$ m $^3$), imaging the space between the males’ releasing point downstream and male landing point on the odor source of female cages upstream. 

Trajectory data of successful flights (i.e., males reaching the odor source) were processed and calibrated. Two flow conditions were chosen to test the effect of the wind flow velocity disturbances on the male moth’s flight trajectory, called undisturbed and disturbed. Undisturbed refers to a relatively steady flow maintained in the wind tunnel due to mesh screens at the wind tunnel’s inlet. Disturbed corresponds to an additional vertically oriented cylinder of 0.05 m in diameter and 1.0 m high, positioned 0.2 m upstream of the female cages. We ensured that the odor source was not in the cylinder recirculation region. The wake formed by this cylinder is characterized by repeating patterns of swirling vortices downwind from the cylinder, called the von Kármán vortex street.

\subsubsection{Wind tunnel measurements}

\begin{figure}
    \centering
    \includegraphics[width=0.99\textwidth]{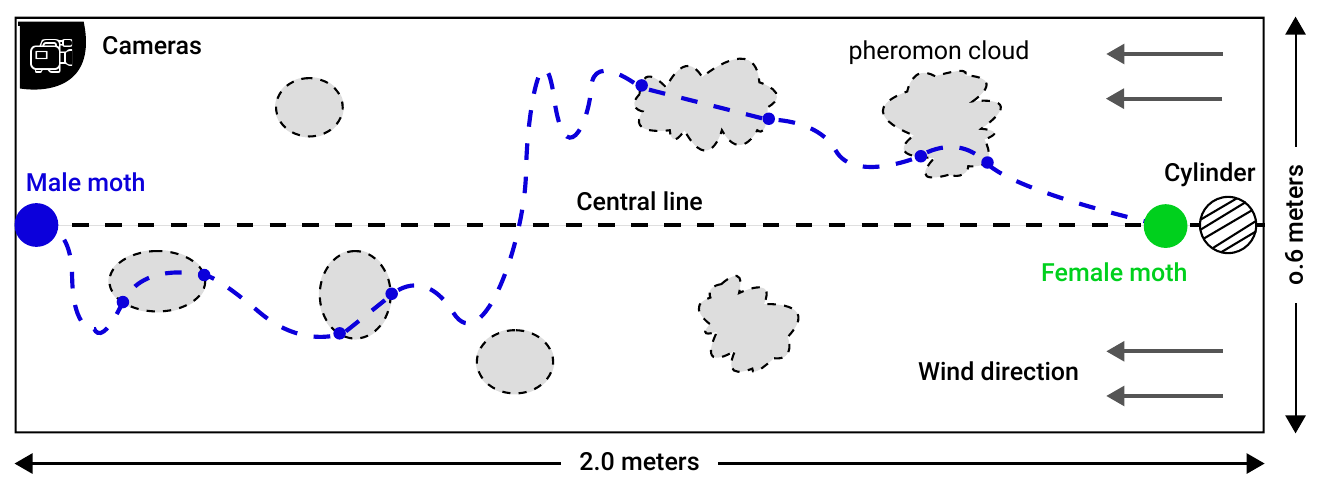}
    \caption{A top-view schematic view of the experimental setup and a schematic picture of navigation flight crossing dispersed pheromone clouds. The moth flights are from left to right, and the wind is from right to left. The navigator’s exploration is responsible for searching for the next pheromone signal, and exploitation is flying upwind after the pheromone is encountered. The exploration/exploitation ratio (EER) varies as the navigator approaches the source due to the smaller average plume width and more rapid concentration variations. Increasing turbulent intensity will affect the dispersion close to the source more dramatically.}
    \label{fig:exp}
\end{figure}

The flight paths of male moths were tested at the peak of calling time, 4-5 hours after the onset of the scotophase, at 25-27$^\circ$ C. The female cage was located 0.2 m downwind from the end of the tunnel at the tunnel center line (0.5 m). Males were randomly taken individually from the male cage and released at the downstream releasing midpoint, 2 m downstream of the odor source cages. Each male was free to take off. The dataset presented here includes 57 successful flights, where 35 experienced disturbed flow while the other 22 took flights in undisturbed flow. Fig.~\ref{fig:exp} gives a top-view schematic view of the experimental setup.

\subsection{Olfactory navigation simulator}

In this study, we adopted the olfactory navigation simulator, MothPy, proposed by Ref.~\cite{Golov:2021}. The navigators are self-propelled flyers with a single sensor that provides binary odor detection (below/above its sensor threshold) and a timer without memory or learning. The plume model is based on another simulation for the plume of pheromone distribution in a turbulent flow~\cite{Farrell2002}. Importantly, to represent spatially local sensing, the simulated agent could only access the local wind velocity and odor detection above the threshold at its position. 

Although MothPy was successfully utilized in various studies, demonstrating the central concepts in moth-inspired navigation algorithms, it could not create paths comparable with measured moth flight trajectories. First, it was designed for two-dimensional plumes, while the trajectory data is captured in three dimensions. Second, the simulator allowed the self-propelled navigation agent to accelerate and maneuver instantly. Finally, the agent’s ability to accurately sense its state in terms of the wind’s direction and intensity as well as the movement of the pheromone clouds. We first extended the simulator to three dimensions, see MothPy 3D.
Moreover, we introduced the acceleration limit to obtain more realistic flight paths. Lastly, we present a parameter that adds Gaussian noise concerning the mean (\(\mu\)) and standard deviation (\(\sigma\)) to the input signals of the agent. This better mimics the sensory error for realistic environments, allowing the algorithm to accurately capture the moth's navigation patterns. 

We conducted the following fitting procedure to evaluate how well the proposed simulator captures empirical data. First, for each measured path, we matched a synthetic simulated path. Second, to obtain the artificial path, we ran the simulator for \(100\) times and took the average route. We computed multiple runs and considered the mean path. This is due to the stochastic nature of the simulator and because we aim to investigate the \say{principal} behavior a moth presents during olfactory navigation. Since the paths are not the same length, we used the dynamic time wrapping method with a mean absolute error (MAE) metric to compare the artificial and experimental paths~\cite{dtw}. In addition, we computed the coefficient of determination between these two to determine the portion of the data \say{explained} by the model. Thus, the outcome of this procedure is a value, \(r \in [0, 1]\), which indicates how well the observed moth navigation paths are replicated by the proposed simulator based on the proportion of total variation of outcomes explained by the simulator.

\subsection{Exploration-exploitation model}

We propose an EE dilemma-based theoretical model to explain moth olfactory navigation. We first apply the model to the MothPy 3D simulated paths. We assume that the agent's path can be divided into \(k \in \mathbb{N}\) ordered subsets (\(P\)) such that each subset (\(p_i \in P\)) indicates either exploration (\(s\)) or exploitation (\(o\)). Two consecutive subsets cannot have the exact identification. Formally, 
\begin{equation}
\forall i \in [1, \dots, k]: p_i = s \leftrightarrow p_{i-1} = p_{i+1} = o \wedge p_i = o \leftrightarrow p_{i-1} = p_{i+1} = s    
\end{equation}

This definition is appropriate as presumably moths start their flight only when they receive some pheromone trigger, which means they have a pre-defined (and unclear) goal from the moment they take off. 

The path is represented as a vector \(l \in \mathbb{R}^{3 \times n}\) of size \(n \in \mathbb{N}\) of real three-dimensional vectors (\(\forall i \in [1, \dots, n]: l_i \in l\)) corresponding to the agent's center of mass. Since the absolute location in space is not informative, we computed the velocity vector at each point in the path using the numerical \textit{five-point stencil} \cite{numerical_scheme} formula \(v_i := (-l_{i-2} + 8l_{i-1} - 8l_{i+1}+l_{i+2})/12\). To measure the levels of exploitation-exploration rate, which quantifies the amount of exploration the agent performs in a given segment in time, of a subset that starts in index \(t_s\) and ends in index \(t_e > t_s\), we adopted the metric proposed by \cite{hagabubim}. Namely, the exploration score of a path \(p\), which is represented by a velocity vector (\(V\)), is defined as follows. 
\begin{equation}
o(p_i) :=  \frac{1}{2 \pi} \sqrt{\frac{\sum_{v_i \in V} (v_i - \bar{v})^2}{t_e - t_s}}
\end{equation}

\noindent where \(\bar{v} := \sum_{v_i \in V} (v_i) / (y-x)\) \cite{metric_1,metric_2}. Intuitively, this metric measures how aligned the vectors are toward a central direction (\(\bar{v}\)), the mean velocity vector in the segment. Following this metric, a division of the path according to the EE model with \(k\) subsets maximize 
\begin{equation}
    \begin{array}{cc}
        \sum_{i \in [0, \dots k]} \begin{cases}
                        o(p_i), \; p_i = o \\
                        -o(p_i), \; p_i = s
                    \end{cases} .
    \end{array}
    \label{eq:metric}
\end{equation} 

In practice, Eq.~\eqref{eq:metric} imposes a non-linear optimization task that can be solved using various methods. We adopted a heuristic approach for this task to balance the computational burden, solution simplicity, and accuracy of the obtained results. First, a \say{guess} of the number of subsets, \(k\), to range between two and arbitrary large number, \(z \in \mathbb{N} < n\). For each value of \(k\), one must pick \(k-1\) indexes for the division aiming to maximize Eq.~\eqref{eq:metric}. We used a genetic algorithm (GA) approach \cite{ga_1,ga_2,ga_3,ga_4}. First, a random \say{population} of size \(\alpha \gg 1 \in \mathbb{N}\) where each individual in the population is defined by the \(k-1\) indexes, chosen randomly uniformly between the values \(x\) and \(y\). Following the standard practice for GAs, we expressed three operators: mutation, cross-over, and next generation. For the mutation operator, an index is chosen randomly and either increased or decreased by a value of one with a 0.5 probability; the cross-over operator consists of two individuals - \(I_1\) and \(I_2\) starts by randomly choosing an index \(j \in [2, \dots, k-2]\) uniformly. Based on this index, two new individuals are created as follows: 
\begin{equation}
I_1^{new} := I_1[1, \dots j] \cup I_2[j+1, \dots k-1] \wedge I_2^{new} := I_2[1, \dots j] \cup I_1[j+1, \dots k-1]
\end{equation} 

Finally, we adopted the \say{tournament with royalty} operator for the new-generation operator. Initially, the fitness of all individuals in the population is computed using Eq. (\ref{eq:metric}) and ordered from high to low. A portion of the population, \(\beta \in (0, 1)\), is kept for the next generation. In addition, other individuals in the population are selected using the following tournament selection process. Suppose there are \(w\) individuals to be chosen. In each iteration, an individual is selected from the remaining population with a probability corresponding to its normalized fitness score (normalized such that the sum of all individuals' fitness scores is \(1\)). This process is repeated until \(W\) individuals are selected. The three operators are repeated for a pre-defined \(\zeta \in \mathbb{N}\) number of times. At the end, the individual with the highest fitness score is taken. At this stage, we have \(z-1\) possible EE divisions corresponding to different values of \(k\). Therefore, we use the elbow-point \cite{elbow,elbow_2} to find the best value of \(k\) and the final EE division. 

As EE divisions are known, we used the simulator to generate many paths while also recalling the steps in which the simulator altered its state to be the baseline of the EE division. Using these settings, we evaluate the accuracy of the proposed EE division method by computing the precision of a binary classification task. Namely, the EE division classifies each step in time into either \(o\) or \(s\). If the division is done perfectly, the accuracy score would be one, while errors in the division location or number of divisions would lower this score. 

\subsection{Moth-inspired navigation strategy analysis}

The pheromone source location and its quality play a crucial role in the moth's decision-making process during navigation. We compared this strategy with other published navigation algorithms~\cite{Golov:2021}, demonstrating that the navigator based on the idea of exploitation rate proportional to the previously acquired pheromone signal yields higher success rates. An essential part of this strategy relies on wind direction and turbulent fluctuations, determining the mixing and spreading rates of the pheromone clouds. 

Since turbulent and disturbed flows vigorously mix and disperse the pheromone clouds, as shown schematically by the ``envelope'' of the pheromone plume, this should result in more exploration efforts for higher flow velocity fluctuations. Thus, we hypothesize that there is a monotonic decreasing connection between the distance from the target (the female moth) and the exploration-exploitation rate. To compute the exploration-exploitation rate, we first EE divide the path of each male moth. Right after, we calculated the exploration-to-exploitation rate (EER) in locations with some Euclidean distance from the target, up to a 0.06 m error radius, and reported the mean and standard deviation values. We divide the cases where the cylinder was and was not present to account for the level of disturbance to the mean flow. 

\section{Results}
\label{sec:results}

In this section, we present the results obtained from the experiments, divided into the simulator's ability to capture and explain real-world data, how accurate the exploration-exploitation model and its fitting method, and the connection between the distance from the target and exploration requirements for different turbulence settings. 

\subsection{Simulator efficiency}

Table \ref{table:simulator_fitting} shows the mean absolute error (MAE) and the coefficient of determination (\(R^2\)) of the simulator compared to the real-world data, using the different subsets of the three improvements added to the MothPy simulator. With all three modifications included, the simulator can explain 72\% of the data, achieving a 12.7\%  improvement over the previous MothPy version. An additional comparison with the original MothPy with different axes trajectories is provided in the Appendix.

\begin{table}[!ht]
\centering
\begin{tabular}{l|cccc}
\textbf{Configuration} & MothPy-3D & \begin{tabular}[c]{@{}l@{}}MothPy-3D \(+\) \\ bounded accaloration\end{tabular} & \begin{tabular}[c]{@{}l@{}}MothPy-3D \(+\) \\ noisy input\end{tabular} & \begin{tabular}[c]{@{}l@{}}MothPy-3D \(+\)\\ bounded acceleration \(+\)\\ noisy input\end{tabular} \\ \hline
\textbf{MAE} & \(13.845 \pm 2.701\) & \(12.893 \pm 2.948\) & \(13.317 \pm 3.086\) & \(12.748 \pm 3.027\) \\
\textbf{\(\boldsymbol{R^2}\)} & \(0.593 \pm 0.062\) & \(0.712 \pm 0.053\) & \(0.695 \pm 0.064\) & \(0.720 \pm 0.049\) \\ 
\end{tabular}
\caption{The simulator's results fit of the measured flights. }
\label{table:simulator_fitting}
\end{table}

\subsection{Exploration-exploitation model evaluation}
\begin{figure}
    \centering
    \includegraphics[width=0.99\textwidth]{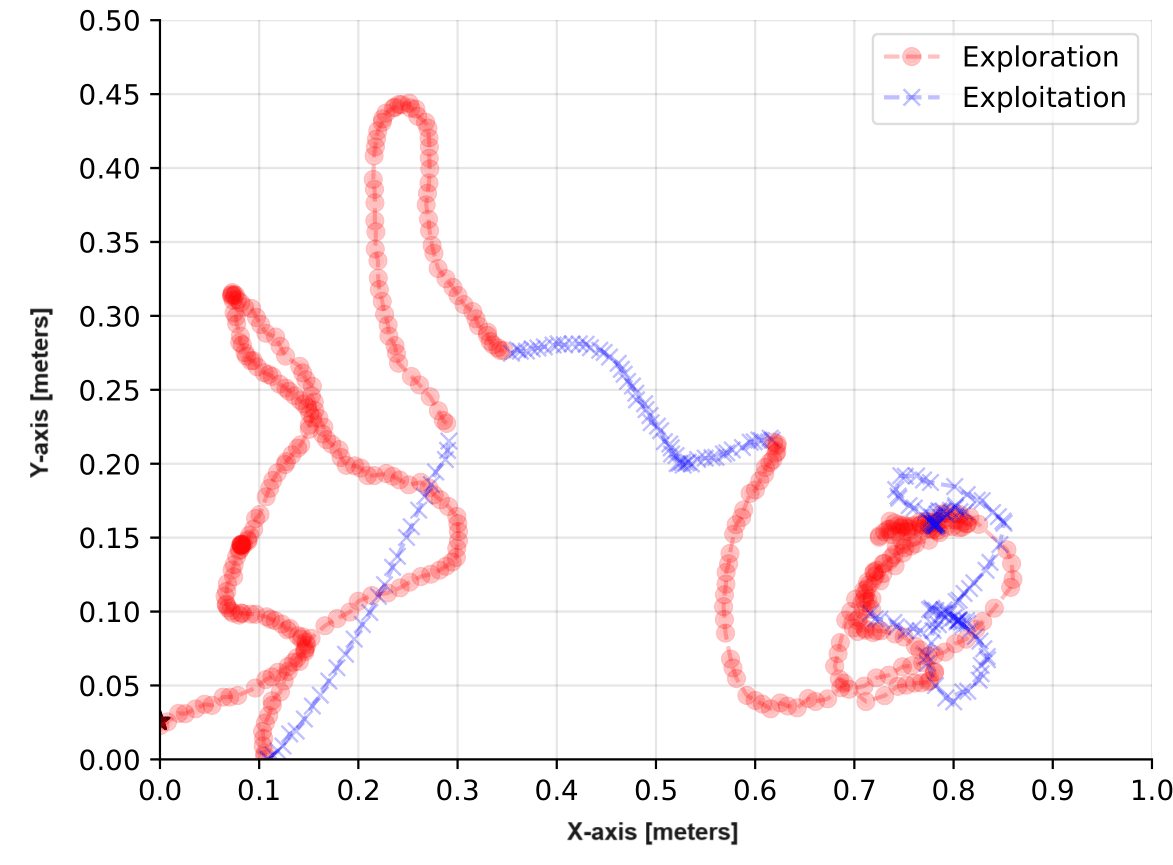}
    \caption{An example of the exploration-exploitation split of a single moth flight. The exploration and exploitation phases are indicated by (red) circles and (blue) axes. The color intensity is correlated to the number of times the agent is located in the same (\(x,y\)) coordinates with different or identical \(z\) coordinate values. }
    \label{fig:ee-example}
\end{figure}

To evaluate the EE model, the simulator generated 10000 paths, balancing between computational burden and a large sample to obtain statistically significant results. For each track, the number and the indexes of the divisions between the exploration and exploitation phases are recalled. The proposed EE fitting procedure is then computed on these paths, identifying each step for each route as either exploration (\(s\)) or exploitation (\(o\)). Overall, we obtained an accuracy score of 84.46\%. For example, Fig. \ref{fig:ee-example} shows the random exploration-exploitation split of a single moth flight, as computed by the algorithm. 

\subsection{Supporting our EE model hypothesis using disturbed flow case}

\begin{figure}
    \centering
    \includegraphics[width=0.99\textwidth]{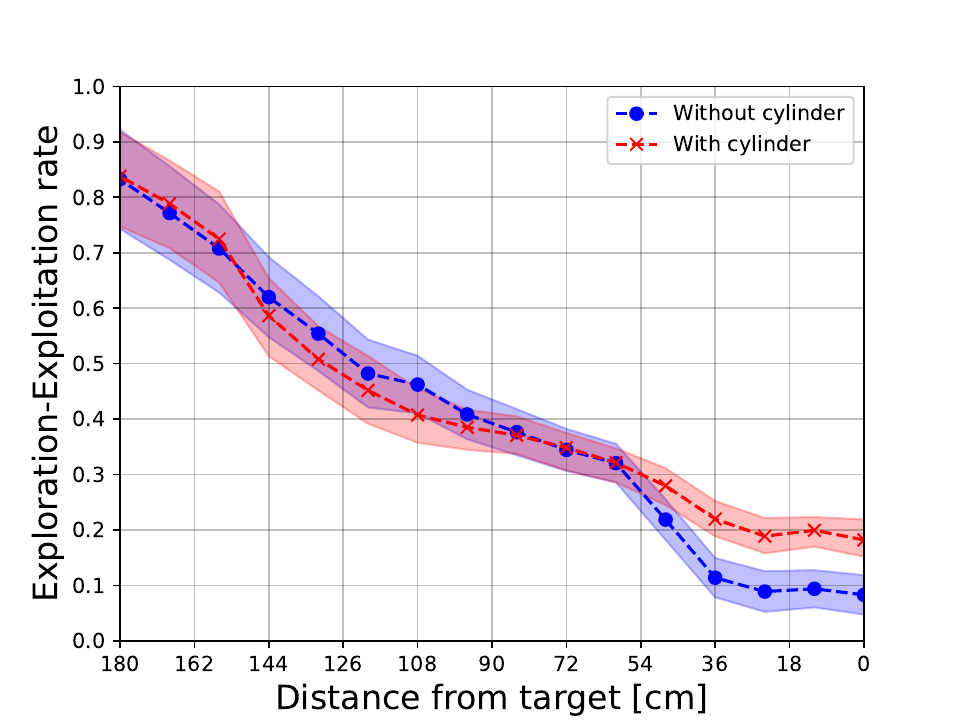}
    \caption{A comparison between the cylinder and no-cylinder experiments regarding the exploration-exploitation rate as a function of the distance from the agent's target. The curve is the mean, and the shadowed bound is the standard deviation of \(n=35\) and \(n=22\), respectively.}
    \label{fig:alex}
\end{figure}

Fig. \ref{fig:alex} depicts the flight strategy path where the x-axis indicates the Euclidian distance of the agent (i.e., the male moth) from the target (the female moth) in centimeters. The y-axis indicates the exploration-exploitation rate. The results are shown as each set's mean \(\pm\) standard deviation as a color-shaded envelope. In addition, we divided the cases for undisturbed and disturbed flows, indicated by the red axes and the blue circles, respectively. As hypothesized, there is a monotonic decrease in exploration to exploitation rate concerning the distance from the target. One can notice that both cases behave similarly up to 0.6 meters from the target. However, at smaller distances, the two cases separate such that the case with the cylinder present decreases more slowly than the other, resulting in relatively higher EER. 

\begin{figure}
    \centering
    \includegraphics[width=0.9\textwidth]{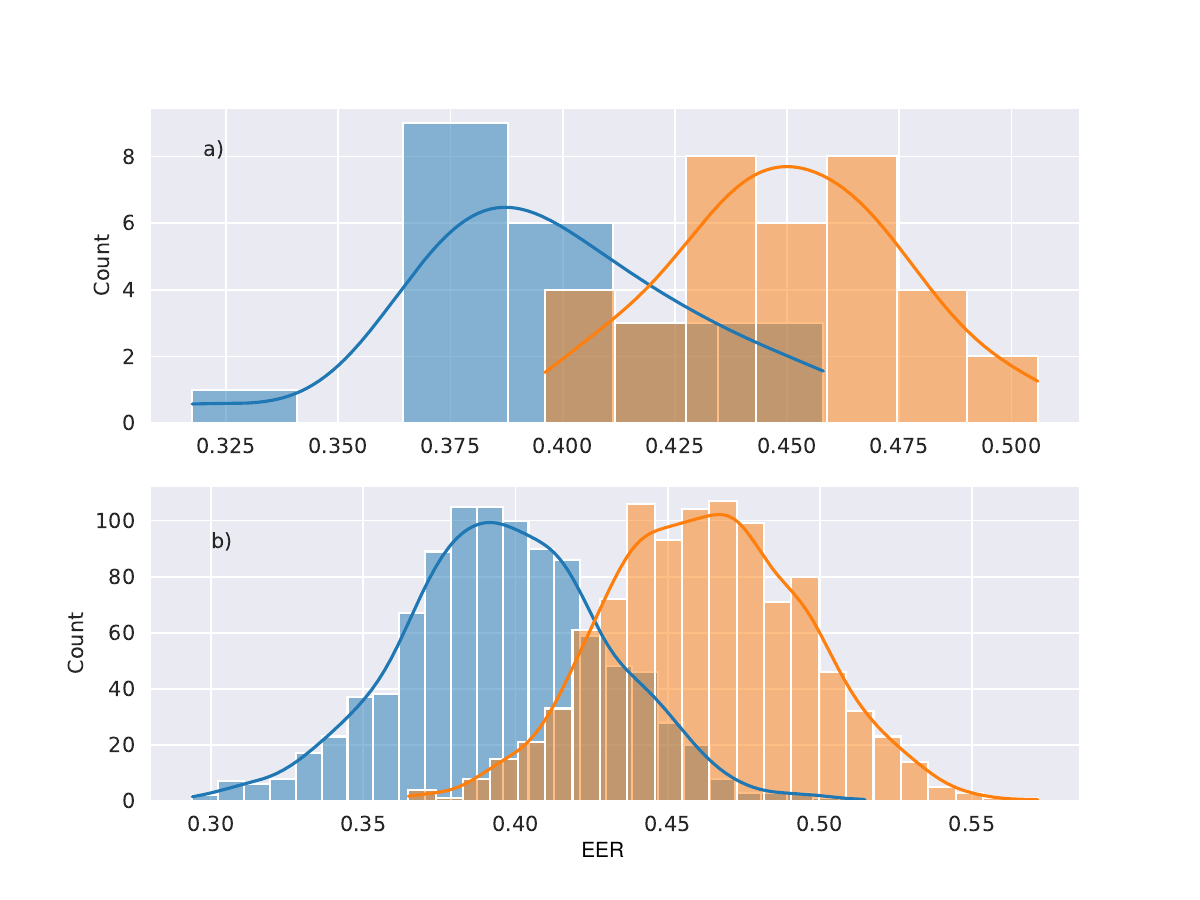}
    \caption{Histograms for the undisturbed vs. disturbed flow experiments (a) and simulations (b), presenting the average exploration-exploitation rate (EER). The left (blue) plot is without a cylinder (undisturbed), and the right (brown) plot has increased flow disturbance due to the cylinder (disturbed flow experiments) or due to the increased turbulent intensity parameter (turbulent-like flow simulations).}
    \label{fig:alex2}
\end{figure}

Fig. \ref{fig:alex2} presents the histograms of the cylinder (red) and no cylinder (blue) experiments depicting the average exploration-exploitation rate such that the x-axis indicates the average exploration-exploitation rate of each flight. The y-axis indicates the number of flights with the same EER. The lines show a kernel density approximation of the data, added for clarity. As suggested, there is a monotonic decrease in exploration to exploitation rate concerning the distance from the target. Noticeably, in the cases where the cylinder is present, the exploration-exploitation rate is higher, with almost 5\% more, on average. Furthermore, the cases in the undisturbed flow have less diverse dynamics than those in the disturbed flow. 

\section{Discussion and conclusions}
\label{sec:discussion}

The exploration-exploitation (EE) dilemma is a fundamental concept in decision-making that has implications across various domains \cite{ee_human_2,ee_human_3,eve_human}. This paper shows that an EE-based model can explain moths' olfactory-driven flights. We offer a memory-free algorithm that captures only the upwind direction and binary indication for the presence of pheromones - both with some level of noise. 

We improved the moth flight simulation, MothPy, by introducing three modifications: three-dimensional settings, bounded acceleration vector size, and Gaussian noise in the sensory input. As shown by Table \ref{table:simulator_fitting}, with these modifications, the proposed simulator obtains a coefficient of determination of \(R^2 = 0.720\) when tested on \(n=57\) moth flight paths. For comparison, the simulator received only \(R^2=0.593\) for the same data without the changes, indicating a 21.4\% relative and 12.7\% absolute improvement. Based on these results, it is suggested that the simulator can capture the central navigation dynamics of a moth while missing local and settings-dependent conditions such as the influence of temperature, humidity, and the agent's properties (such as size) \cite{Golov:2022}. 

Using the proposed simulator, we show that the EE model can explain the moth' olfactory navigation. Indeed, when fitting an EE model on the simulator data, which describes 72\% of the real-world data, we obtained that 84.5\% of the time, the EE model predicted accurately when the agents utilize a known signal and when it searches for such signal. This outcome is unsurprising as EE is known to explain many animal behavior types in nature \cite{ee_dis_1,ee_dis_2,ee_dis_3}. From a computational point of view, EE is useful as a model for optimization and search tasks \cite{ee_dis_4,ee_dis_5}.  

Since the EE model can reasonably explain the moth's olfactory navigation, one can use it to describe the changes in the flight behavior of the moth in different settings. To this end, we focused on the effect turbulence causes. Thus, the main result is in Fig.~\ref{fig:alex}, as it demonstrates that the original hypothesis of the moth-inspired navigation algorithm \cite{Liberzon2017} highlighted here in terms of EE model, explains the observed moth flight paths: a) exploration-exploitation rate reduces as the male approaches female source due to the plume width as a function of distance, or exploration is proportional to the plume width b) turbulence affects pheromone plume dispersion rate in the proximity to the female source, and it increases the exploration rate. Thus, the search efforts are reduced due to the distance from the target. Indeed, shown in Fig. \ref{fig:alex} that for both the cases with flow disturbance (i.e., the cylinder is present) and without it, for distances higher than 0.6 m, the exploration-exploitation rate decreases linearly. After this point, the presence of a cylinder, which potentially increases the turbulence, causes the agents to spend almost twice the exploration efforts compared to the case where it flows virtually in parallel and does not cause much distortion. Fig. \ref{fig:alex2} further supports this outcome as the mean exploration-exploitation rate for the cases where the cylinder is present is higher, on average, and causes more diverse navigation behavior.  

Generally, when considering the classical diffusion dynamics from a computational perspective, the \say{information} it caries within the concentration field reduces exponentially as a function of the distance from the source. Hence, an agent would have to spend more time on random searches to collect enough information, which is decaying and spreading across a larger volume of space. Moreover, turbulence or disturbed flow have super-diffusion properties~\cite{diff}. Hence, the diffusion dynamics is a lower-boundary to the search required for an agent. The results are unsurprising when treating the dynamics as a spatio-temporal optimization task based on an agent's available information.

This research has several limitations. First, the biological data used as a baseline was obtained in sterile settings where only a single goal is present to the male moth, and environmental factors such as temperature and humidity were fixed. As such, the used data only approximates what realistically happens in nature. Future work can focus on curating and using such data to evaluate the proposed simulator and model. Second, the proposed simulator assumes moths have only a single pheromone sensor. However, moths can differ between multiple pheromone types and choose conspecific females~\cite{Golov:2022}. Ignoring this intricate sensory system could limit the model's accuracy and ability to explain the moth navigation. Exploring different sets of pheromone senses and the computational effort they require is a promising future venue. Third, while we show a stochastic EE model to explain the data, it is fair to assume that a deterministic but more complex model would be able to determine the phase changing between exploration and exploration, which can benefit engineering developments based on the proposed model. Finally, there are two different diffusion coefficients for vertical and horizontal directions \cite{Liberzon2017}, but we treated them identically, which can result in less accurate physical simulation in near-ground settings.

Our results highlight the potential of using the well-established EE model alongside moth-inspired olfactory navigation to obtain an agent based on a single spatially sparse signal with noise due to turbulence. This outcome can be the foundation of robotic agents to solve similar navigation tasks. 

\section*{Declarations}
\subsection*{Funding}

This research received no specific grant from funding agencies in the public, commercial, or not-for-profit sectors.

\subsection*{Conflicts of interest/Competing interests}
None.

\subsection*{Data availability}
The data used in this study is available by a formal request from the authors. 

\subsection*{Author contribution}
Teddy Lazebnik: Conceptualization, Methodology, Software, Formal analysis, Investigation, Visualization, Project administration, Writing - Original Draft, Writing - Review \& Editing. \\ Yiftach Golov: Data curation. \\ Roi Gurka: Data curation, Writing - Review \& Editing. \\ Ally Harari: Data curation. \\ Alex Liberzon: Conceptualization, Methodology, Formal analysis, Investigation, Validation, Writing - Review \& Editing. \\ 
 
\bibliography{biblio}
\bibliographystyle{unsrt}

\clearpage
\section*{Appendix}

Intuitively, one can compare the 3D real-world data with a 2D simulator by projecting the 3D data into the XY, XZ, or YZ two-dimensional plans and conducting the same analysis. Table \ref{table:simulator_fitting_2d} compares the 2D version of the MothPy simulator in all possible projection comparison settings to the MothPy-3D version. One can notice that the XY is better compared to the XZ and YZ trajectories, while all three are slightly worse, on average, compared to the 3D version of the simulator. Moreover, we used a non-parametric k-sample Anderson-Darling test \cite{stat_test} to determine if the cases are statistically significantly different from one another with \(p < 0.05\). The test reveals that while the 3D case outperforms the 2D cases with XZ and YZ trajectories, it does not surpass the XY trajectory (\(p = 0.094 > 0.05\)). Interestingly, when computing the contribution of each axis using a principal component analysis (PCA) analysis, one obtains that the X, Y, and Z axes are capturing 42\%, 35\%, and 23\% of the dynamic's information, respectively. This outcome can explain why the XY trajectory outperforms the other two trajectories as it relatively captures more data. Notably, unlike Table~\ref{table:simulator_fitting} that reported the MAE of the simulator, in this experiment, we do not notify the MAE as it is lower for 2D compared to 3D only due to introducing an additional dimension. 

\begin{table}[!ht]
\centering
\begin{tabular}{l|cccc}
\textbf{Configuration} & MothPy-3D& MothPy-2D XY & MothPy-2D XZ & MothPy-2D YZ \\ \hline
\textbf{\(\boldsymbol{R^2}\)} & \(0.593 \pm 0.062\) & \(0.570 \pm 0.057\) & \(0.548 \pm 0.044\) & \(0.539 \pm 0.059\) \\ 
\end{tabular}
\caption{Comparison of the 2D version of the MothPy simulator in all possible projection comparison settings to the MothPy-3D version. }
\label{table:simulator_fitting_2d}
\end{table}

\end{document}